\documentclass{ifacconf}

\usepackage{graphicx}      
\usepackage{natbib}        
\usepackage{algorithm}
\usepackage{algpseudocode}

\usepackage{amsmath,amsfonts}
\usepackage{subcaption}
\usepackage{array}
\usepackage{textcomp}
\usepackage{balance}
\begin{document}
\begin{frontmatter}

\title{Probabilistic RRT Connect with intermediate goal selection for online planning of autonomous vehicles} 


\author[First]{Darshit Patel}
\author[First]{Azim Eskandarian}

\address[First]{Department of Mechanical Engineering, Virginia Tech, Blacksburg, VA 24061 USA (email: darsh2198@vt.edu, eskandarian@vt.edu)}

\thanks{This work has been submitted to IFAC for possible publication}

\begin{abstract}                
Rapidly Exploring Random Trees (RRT) is one of the most widely used algorithms for motion planning in the field of robotics. To reduce the exploration time, RRT-Connect was introduced where two trees are simultaneously formed and eventually connected. Probabilistic RRT used the concept of position probability map to introduce goal biasing for faster convergence. In this paper, we propose a modified method to combine the pRRT and RRT-Connect techniques and obtain a feasible trajectory around the obstacles quickly. Instead of forming a single tree from the start point to the destination point, intermediate goal points are selected around the obstacles. Multiple trees are formed to connect the start, destination, and intermediate goal points. These partial trees are eventually connected to form an overall safe path around the obstacles. The obtained path is tracked using an MPC + Stanley controller which results in a trajectory with control commands at each time step. The trajectories generated by the proposed methods are more optimal and in accordance with human intuition. The algorithm is compared with the standard RRT and pRRT for studying its relative performance.
\end{abstract}


\end{frontmatter}

\section{Introduction}
Online motion planning is crucial in enabling safe and efficient autonomous driving. As self-driving vehicles navigate complex and dynamic environments, online motion planning algorithms continuously compute feasible trajectories in real-time, allowing the vehicle to respond to changing road conditions, traffic, and obstacles. A thorough survey on the current techniques pertaining to motion planning was done by \cite{MPReview}. It categorizes the planning algorithms into traditional methods and learning-based methods.

Sampling-based motion planning is a subcategory of traditional methods. In these algorithms, a tree is formed from the start point to the destination point by randomly sampling points in the environment. Rapidly Exploring Random Trees (RRT), \cite{RRT}, and Probabilistic Road Maps (PRM), \cite{PRM}, are amongst the earliest and standard techniques in this category. Various modifications to the standard algorithms have been introduced to increase their efficiency and optimality \cite{SampReview}. RRT and its variants are very simple to implement and have been proven to be more effective in dynamic environments than PRM-based methods.

Due to the uniform sampling property of RRT, the convergence time is high for autonomous driving scenarios as it explores the unnecessary regions as well. Hence, RRT is often used for high-level planning instead of online planning. NC-RRT was developed by \cite{NCRRT} to discard unnecessary and far away nodes. They proposed to sample new nodes within a radius of the leaf nodes and increase the radius gradually if a collision is detected. This method can seldom get trapped in local optima and the computational complexity is much higher than the standard RRT resulting in higher convergence time.

In order to reduce the convergence time, Bi-RRT was introduced by \cite{RRTConnect}. In this method, two trees are simultaneously formed from the start and destination points. The final path is generated when any branch of the two trees intersects each other. A small change in the environment results in the trees being broken and replanning is initiated. Dynamic Domain RRT (DD-RRT) was introduced by \cite{DDRRT} where the environment is divided into subdomains. When a change in the environment occurs, replanning is initiated only in that subdomain where the endpoints of the broken tree are considered as start and goal points. 

These techniques often result in discontinuous paths and might not be possible for an autonomous car to follow due to vehicle dynamics constraints. These techniques are usually used for robot motion planning. Closed loop RRT was introduced by \cite{CLRRT} to generate feasible trajectories using the vehicle model. In this method, instead of directly connecting the sampled node to the nearest node, a control command is found using the forward dynamics of the vehicle to advance the trajectory toward the sampled node.

RRT* was introduced by \cite{RRTstar} to obtain an optimal trajectory. ST-RRT* combined the ideas of Bi-RRT and RRT* to obtain an optimal trajectory with lowered average convergence time \cite{STRRT}. ST-RRT* introduced the concept of an intermediate goal region where intermediate goal points are sampled for the goal tree. This algorithm is much faster than the RRT* algorithm for optimality, but the convergence time is still high and is not suitable for online replanning.

\cite{Wu2021} introduced the idea of non-uniform sampling by using a probability density map. The probability of sampling points near the goal is higher and the probability near the obstacles is lower. This increases the sampling efficiency by reducing the sampling in unnecessary regions. They showed that pRRT with high goal bias can be useful for online planning in intersections. Using a high bias value is not favorable for complex environments and lowering the bias value results in higher convergence time which is not favorable for online replanning.

Hence, in this paper, we propose a method to use a highly goal-biased pRRT with additional steps in complex environments to find feasible trajectories around obstacles efficiently. For autonomous driving scenarios, it might be redundant to consider all the obstacles in the surrounding. For example, if a vehicle is making a left turn at an intersection, obstacles on the right side might not be of importance. Hence, we propose a method to recursively select intermediate goal positions around the hindering obstacles and form multiple trees between the start point, destination point, and intermediate goal points. A kinematic bicycle model is used to generate dynamically feasible trees. The multiple trees formed are connected to each other to generate a final trajectory. The performance of the method is then compared to the performance of the standard RRT and pRRT algorithms.

\section{Methodology}
\subsection{Problem Statement}
Given a coordinate space $\mathcal{X}$, an initial position $x_i$, and a destination position $x_d$, we need to find a set of control inputs $u$ such that the ego vehicle moves from $x_i$ to $x_d$ while following the dynamic model $\dot{x}=f(x(t),u(t))$. Here $x(t) \subset \mathcal{R}^4$ represents the state of the ego vehicle and $u \subset \mathcal{R}^2$ represents the control input to the ego vehicle at time $t$. The state $x$ contains the x-y coordinates, $p_x$ and $p_y$, heading angle $\psi$, and the vehicle speed $v$. The control input comprises the steering input $\delta$ and the speed input $v_i$. The coordinate space, $\mathcal{X}$, is a union of two disjoint sets $\mathcal{X}_{free}$ and $\mathcal{X}_{occ}$, where the former represents the free explorable space while the later represents the space occupied by the obstacles. The final generated trajectory should belong to $\mathcal{X}_{free}$ avoiding states in $\mathcal{X}_{occ}$.

\subsection{Probabilistic RRT (PRRT)}
The Probabilistic Rapidly Exploring Random Trees is a modified version of RRT where a position probability map (PPM) is used for the random point sampling (\cite{Wu2021}). A tree is initialized at the start point and it eventually explores the free space and finds a feasible path to the goal point. The function $generatePDFMap()$ creates a probability map that increases the probability density at the goal position and reduces it in the obstacle space $\mathcal{X}_{occ}$. The bias value $\lambda$ defines the scale of the probability change towards the goal and the obstacles. The value of 0 indicates a uniform probability throughout the environment and essentially works like the standard RRT algorithm. $\sigma^2$ parameter indicates the variance of the probability density at the goal. 
\begin{algorithm}
\caption{pRRT}
\label{alg:cap}
\begin{algorithmic}
\State $T \gets InitializeTree(x_i)$
\State $\lambda \gets biasFactor$
\State $PPM \gets generatePDFMap(x_d,\lambda, \mathcal{X}_{free}, \mathcal{X}_{occ})$
\State $i=0$
\While{$i \leq N$}
    \State $x_{rand} \gets sample(PPM)$
    \State $x_{nearest} \gets getNearestNode(x_{rand}, T)$
    \State $x_{new} \gets x_{nearest} + f(x(t), u(t))\cdot\delta T$
    \If{$dist(x_{new},x_{goal}) \leq r_{goal}$}
        \State \textbf{break}
    \EndIf
    \State $i = i + 1$
\EndWhile
\Return $T$
\end{algorithmic}
\end{algorithm}

The steps of this method are summarized in algorithm \ref{alg:cap}. This algorithm returns a tree as its output that connects the start and destination points. Each node of the tree consists of the state of the ego vehicle as well as the control input. The bias value, $\lambda$ affects the sampling of random points significantly. Using a high value reduces the total convergence time but loses the exploration capability to find alternate paths in case of obstacles. Low-value results in increased convergence time for complex environments.

 \begin{figure}[htbp]
    \centering
      \includegraphics[width=0.4\textwidth]{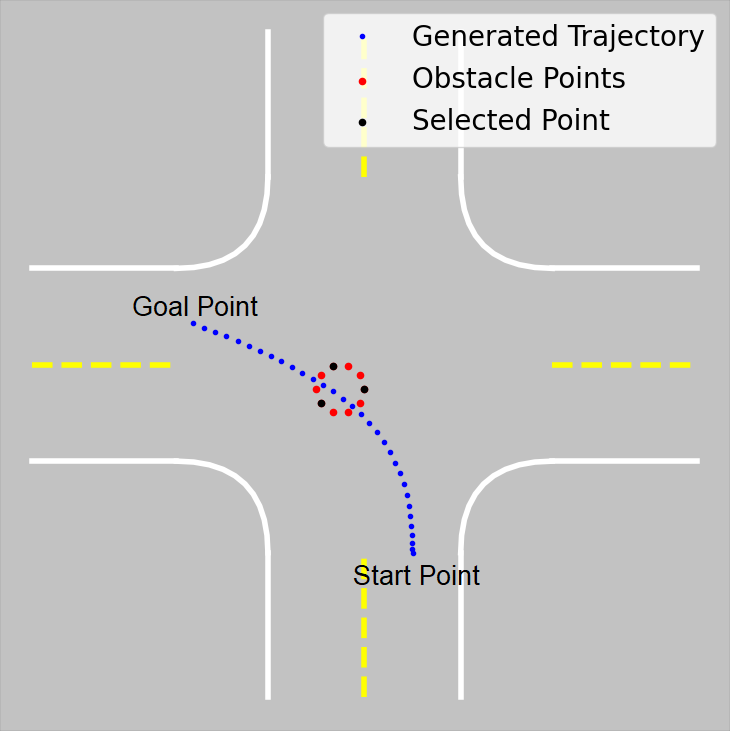}
      \caption{Finding the obstacle points of interest for an obstacle hindering the planned path}
      \label{LeftTurn}
\end{figure}

\subsection{Rapid replanning using intermediate goals}
 We propose extensions to the original pRRT algorithm for quick replanning using high bias values around. In this method, we consider the environment to be free of obstacles. Hence, initially $\mathcal{X}_{occ} = \phi$ and $\mathcal{X}_{free} = \mathcal{X}$. A trajectory is generated from $x_i$ to $x_d$ in this environment using pRRT with a high bias value. A collision check is performed to check if the generated trajectory is hindered by any detected obstacles. Figure \ref{LeftTurn} represents a scenario where a collision is detected in the generated path. 

 \underline{\textit{Obstacle Points of interest:}} The obstacles are discretized into distinct points shown in red color in figure \ref{LeftTurn}. Using all points for intermediate goal selection is redundant and increases the convergence time. Hence only the points with a distance more than half the width of the ego vehicle are selected. This step gives the minimum number of points required for describing the obstacle from the perspective of the ego vehicle. These points are shown in black color in figure \ref{LeftTurn}. 

\begin{figure}[htbp]
    \centering
      \includegraphics[width=0.4\textwidth]{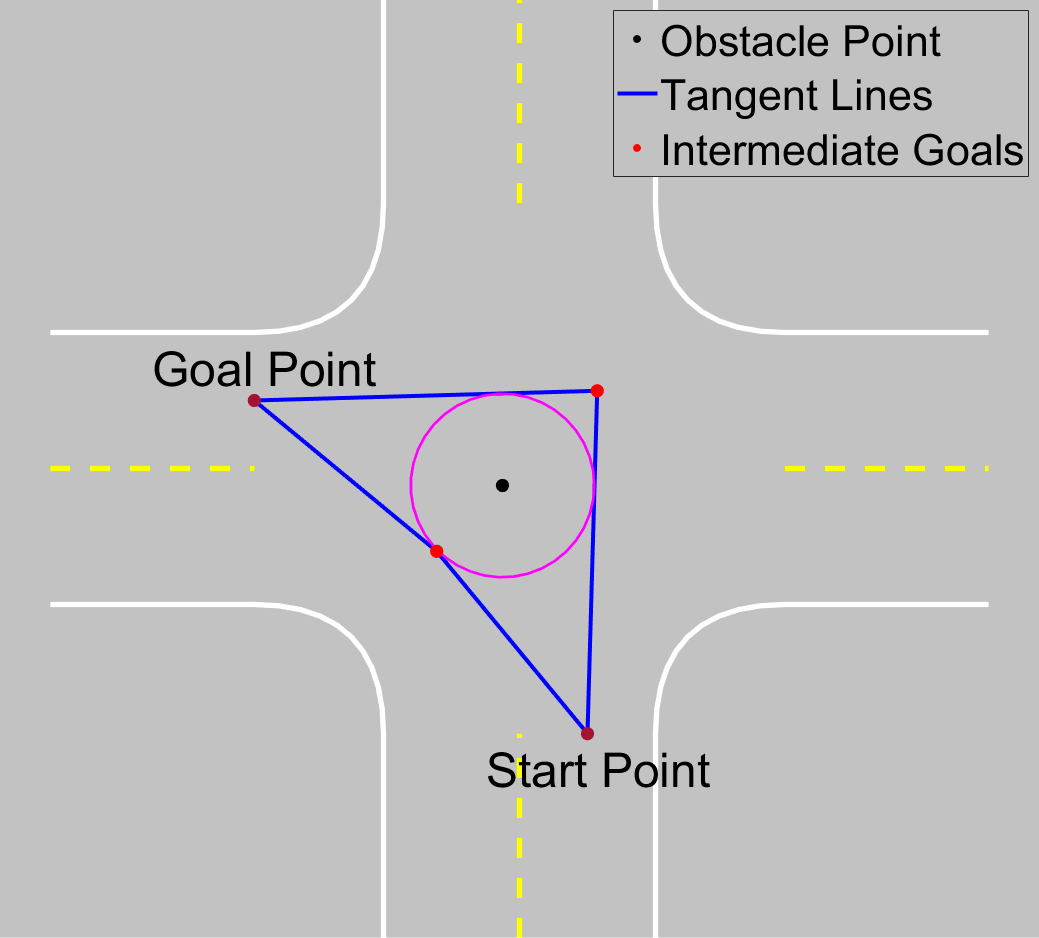}
      \caption{Determining the intermediate goal points around one obstacle point}
      \label{IntGoal}
\end{figure}

\underline{\textit{Get Intermediate Goals:}} For each of the selected points, a safety circle with a radius equal to the width of the ego vehicle is generated. Using one width of the ego vehicle as the radius introduces a safety factor of the half-width of the vehicle for the trajectory. Tangent lines to this circle from the start point and the goal point are determined. The intersection of these tangent lines are selected as the intermediate goal points. This process for a single obstacle point is shown in figure \ref{IntGoal}.

\begin{figure}[htbp]
    \centering
      \includegraphics[width=0.45\textwidth]{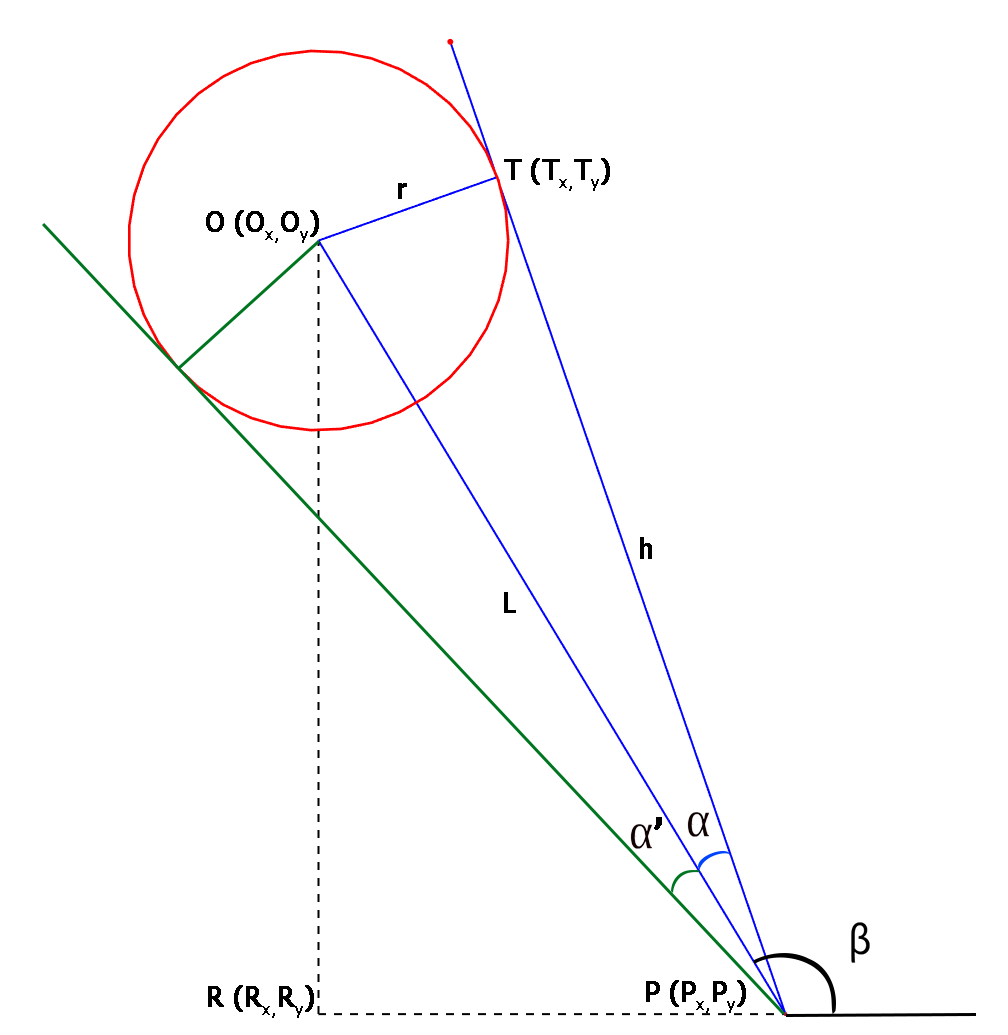}
      \caption{Schematic to find the equation of tangents to the obstacle circle}
      \label{Trigo}
\end{figure}

The intersection of tangents to the circle from two points can be obtained by using simple trigonometry as shown in figure \ref{Trigo}. Considering tangent $\overline{PT}$, the points of intersection between the circle and the tangent can be obtained as follows:
\begin{equation}
\begin{array}{ll}
    & h = \sqrt{L^2 - r^2} \\
    & \beta = atan(O_y-P_y, O_x-P_x) \\
    & \alpha = atan(r,h) \\
    & T_x = P_x + h*cos(\beta-\alpha) \\
    & T_y = P_y + h*sin(\beta-\alpha)
\end{array}
\end{equation}
The equation for the tangent can be obtained using the obtained point $T(T_x,T_y)$ and the start point $P(P_x, P_y)$. Similarly, the equation for the tangent from the goal point can be found. Solving these two linear equations gives us the intermediate goal point. The second goal point is obtained using the other set of tangents. Hence, for each obstacle point, we determine two intermediate goal points.

\underline{\textit{Growing multiple trees:}} Once all the intermediate goal points are found, the pRRT algorithm tries to find feasible trajectories from the start and destination point to the intermediate goal points using a high bias value. The obstacle that hindered the initial trajectory, is now considered in pRRT. This ensures that the new trajectories generated do not collide with the previous obstacles. If for any intermediate goal, pRRT can successfully find two trajectories, a collision check is performed with all the obstacles except for the previously considered obstacle. If a collision is detected, that partial tree is broken down and the algorithm repeats the previous steps with the new start position to be the leaf node of the unbroken tree and destination as the root of the broken tree. The previous steps repeat until all the trees generated are collision-free. Once all the feasible trajectories are generated, the paths obtained from them are connected using a bezier curve.

\underline{\textit{Path Smoothing:}} The final path obtained might be discontinuous and hence the path is smoothened using bezier curves as described by \cite{CAGD}. A bezier curve is a parametric curve defined by the following expression:
\begin{equation}
    P(t)=\sum_{i=0}^n B_i^n(t) \cdot P_i
\end{equation}
$B_i^n$ is known as Bernstein polynomial and its expression is given by:
\begin{equation}
    B_i^n(t) = \binom{n}{i} (1-t)^{n-1}t^i 
\end{equation}
Here $P_i$ are the control points and $n$ is the number of control points. In our problem, we consider the positions of the nodes generated by the two successful trajectories as the control points. The parameter $t$ varies from 0 to 1. Hence, $t\in[0,1]$.

\underline{\textit{Control commands for path tracking:}} The final path obtained by bezier curves only consists of the spatial coordinates for the path. Model Predictive Control (MPC) is used for generating longitudinal control command which is the velocity ($u_2=v$) and a basic Stanley controller is implemented for lateral control command which is the steering angle ($u_1=\delta$).
A timestep of $\Delta t = 0.1s$ was selected for tracking the path. For the MPC controller, a prediction horizon of 20 time-steps and a control horizon of 15 time-steps were selected. The following objective function was used:\\
\begin{equation}
\begin{array} {c c}
    J(v) = &(P-P_{ref})^T \cdot w_1 \cdot (P-P_{ref}) + w_2\cdot Kv^2 \\
    & + (v-v_{max})^T\cdot w_3 \cdot (v-v_{max})
    \end{array}
\end{equation}

Here, $P$ indicates the ego vehicle position, and $P_{ref}$ indicates the reference path. $K$ defines the curvature of the reference path at each location. $v$ is the control variable which is the output of the controller. $v$ is a vector of vehicle speeds over the control period of 15 time-steps. $w_1,w_2,w_3$ are the weights for each minimizing variable.
The steps for the algorithm are summarized in algorithm \ref{alg:PRRTC}.
\begin{algorithm}
\caption{Probabilistic RRT Connect}
\label{alg:PRRTC}
\begin{algorithmic}
\State $HinderingObs \gets \phi$
\State $T \gets PRRT (x_i, x_d)$
\While{$Collision(T,obstacles) = True$}
    \State $HinderingObs.append(obstacle^i)$
    \State $Centers \gets getInterestPoints(obstacle^i)$
    \For{$center in Centers$}
        \State $(x_{g1}, x_{g2}) \gets getIntermediateGoals(center)$
        \State $T_1 \gets PRRT(x_i, x_{g1})$
        \State $T_2 \gets PRRT(x_d, x_{g1})$
        \If{$T_1 \neq \phi$ and $T_2 \neq \phi$}
            \State \textbf{break}
        \EndIf
    \EndFor
    \State $T \gets Connect(T_1, T_2)$
\EndWhile
\end{algorithmic}
\end{algorithm}

\section{Simulation Setup}
\begin{figure}[htbp]
    \centering
      \includegraphics[width=0.35\textwidth]{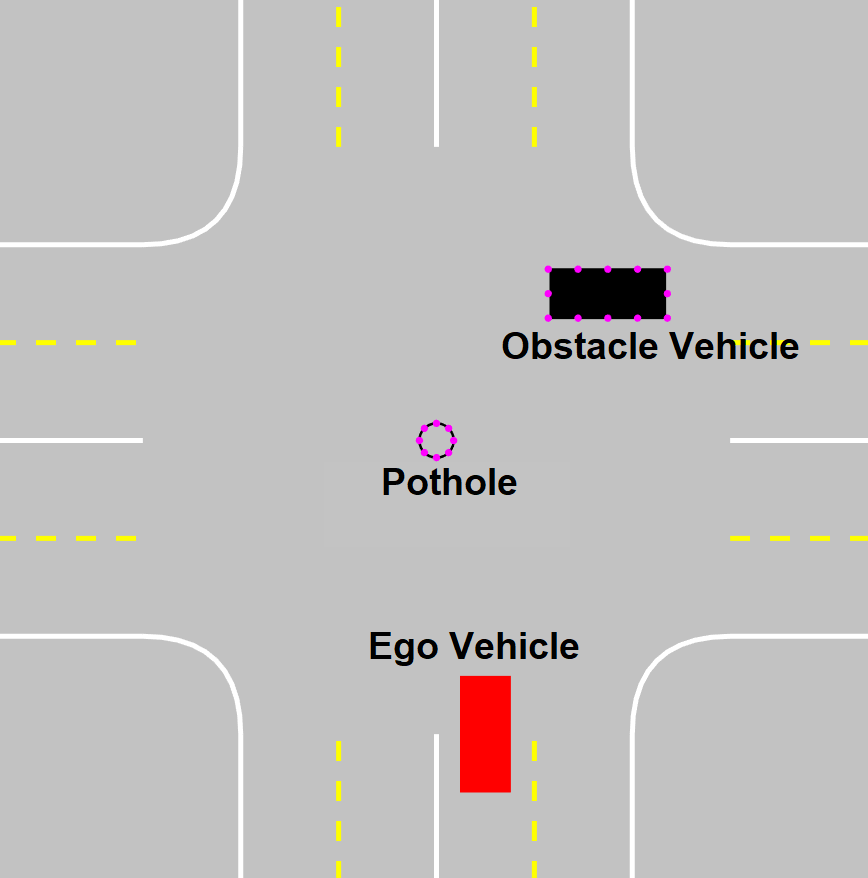}
      \caption{Objects in the environment with their outline points}
      \label{Obstacles}
\end{figure}

The proposed algorithm was implemented in simulation using MATLAB. Various scenarios were tested to validate the feasibility of the algorithm. To test the planner's ability to form trajectories around obstacles. Other vehicles and potholes were introduced as obstacles. Each obstacle was characterized using outline points as shown in figure \ref{Obstacles}.
The MPC was also implemented in order to track the final trajectory obtained by the algorithm. The goal bias value was selected to be $\lambda=10^3$ and the standard deviation was selected as $\sigma = 0.05$. The maximum ego vehicle speed was limited to 10 mph.

\section{Results and Discussion}
\begin{figure}[htbp]
     \centering
     \begin{subfigure}[b]{0.24\textwidth}
         \centering
         \includegraphics[width=\textwidth]{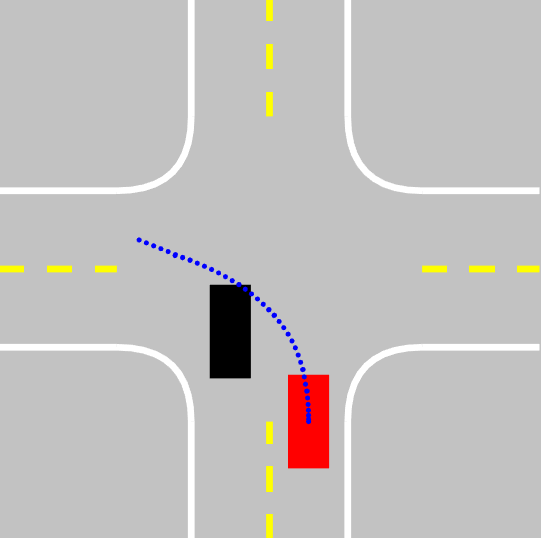}
         \caption{Initial trajectory generated by pRRT}
         \label{fig:1a}
     \end{subfigure}
     \hfill
     \begin{subfigure}[b]{0.24\textwidth}
         \centering
         \includegraphics[width=\textwidth]{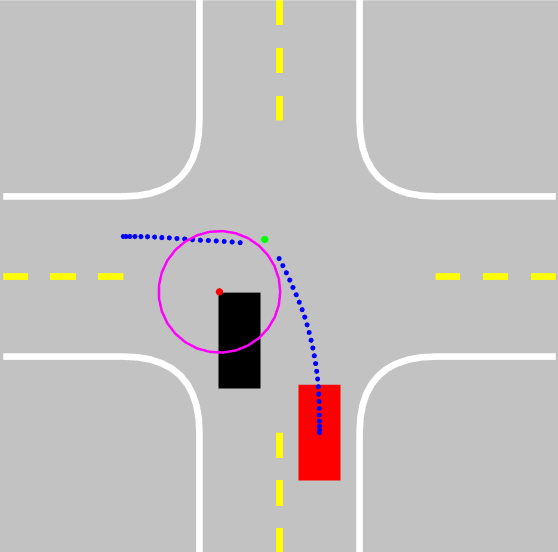}
         \caption{Trees from start and goal towards the intermediate goal}
         \label{fig:1b}
     \end{subfigure}
     \hfill
     \begin{subfigure}[b]{0.3\textwidth}
         \centering
         \includegraphics[width=\textwidth]{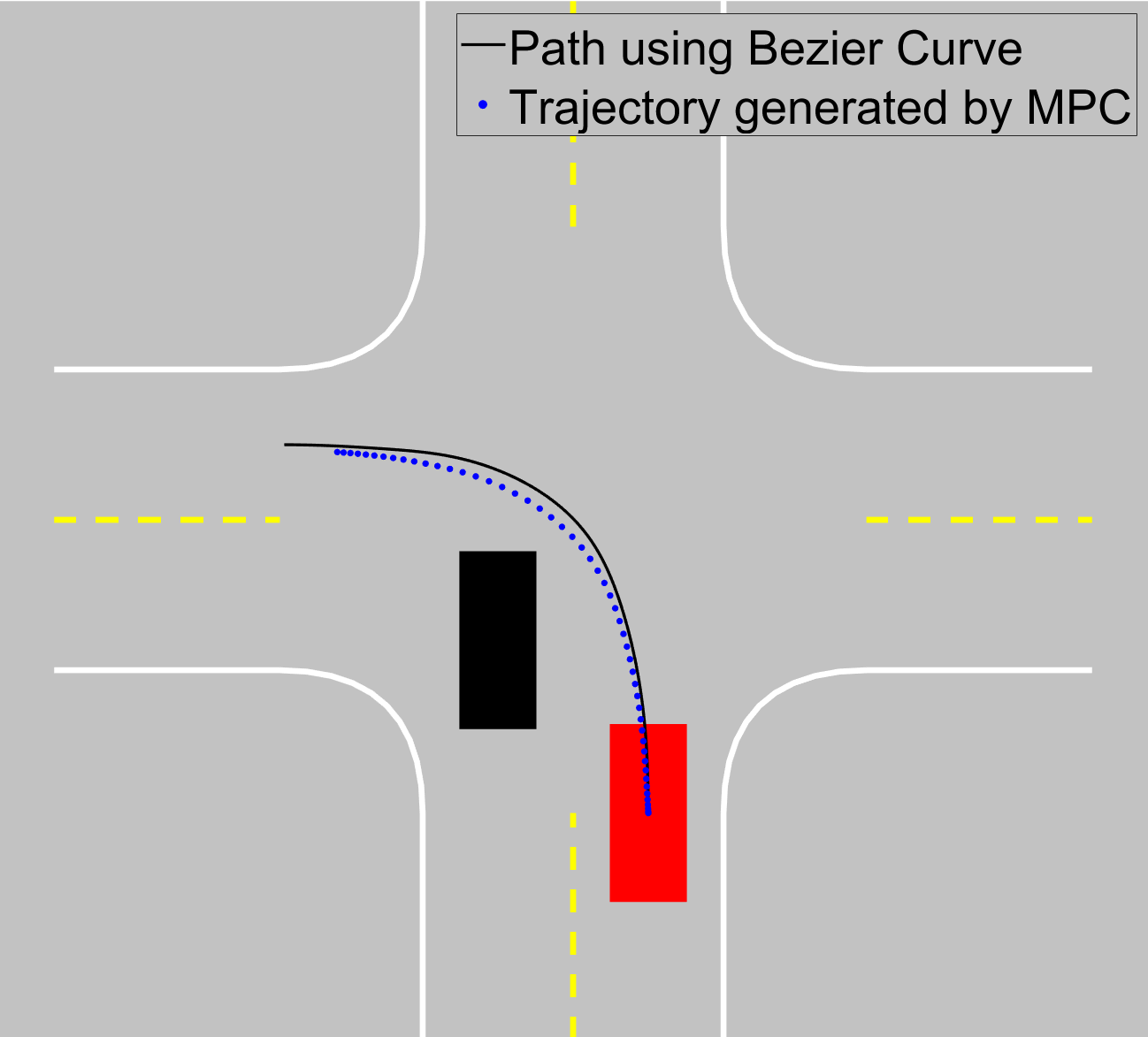}
         \caption{Final trajectory obtained using bezier curve and the generated trees}
         \label{fig:1c}
     \end{subfigure}
        \caption{Left turn maneuver hindered by an obstacle vehicle}
        \label{fig:1LOVHinder}
\end{figure}

Figure \ref{fig:1LOVHinder} shows the case when the ego vehicle tries to make a left turn in a one-lane intersection. Initially, the algorithm considers the environment to be free of obstacles and generates a trajectory using pRRT. Collision check with the detected obstacles show a collision with an obstacle vehicle as shown in figure \ref{fig:1a}. Obstacle points of interest are obtained amongst the points that define the OV as shown in figure \ref{Obstacles}. Intermediate goals for each point of interest as obtained as mentioned in the methodology. Feasible trajectories from the start and goal are obtained to the intermediate goal as shown in figure \ref{fig:1b}. Using the bezier curves, a final path is obtained which is tracked by the MPC controller as shown in figure \ref{fig:1c}.

\begin{figure}[htbp]
     \centering
     \begin{subfigure}[b]{0.24\textwidth}
         \centering
         \includegraphics[width=\textwidth]{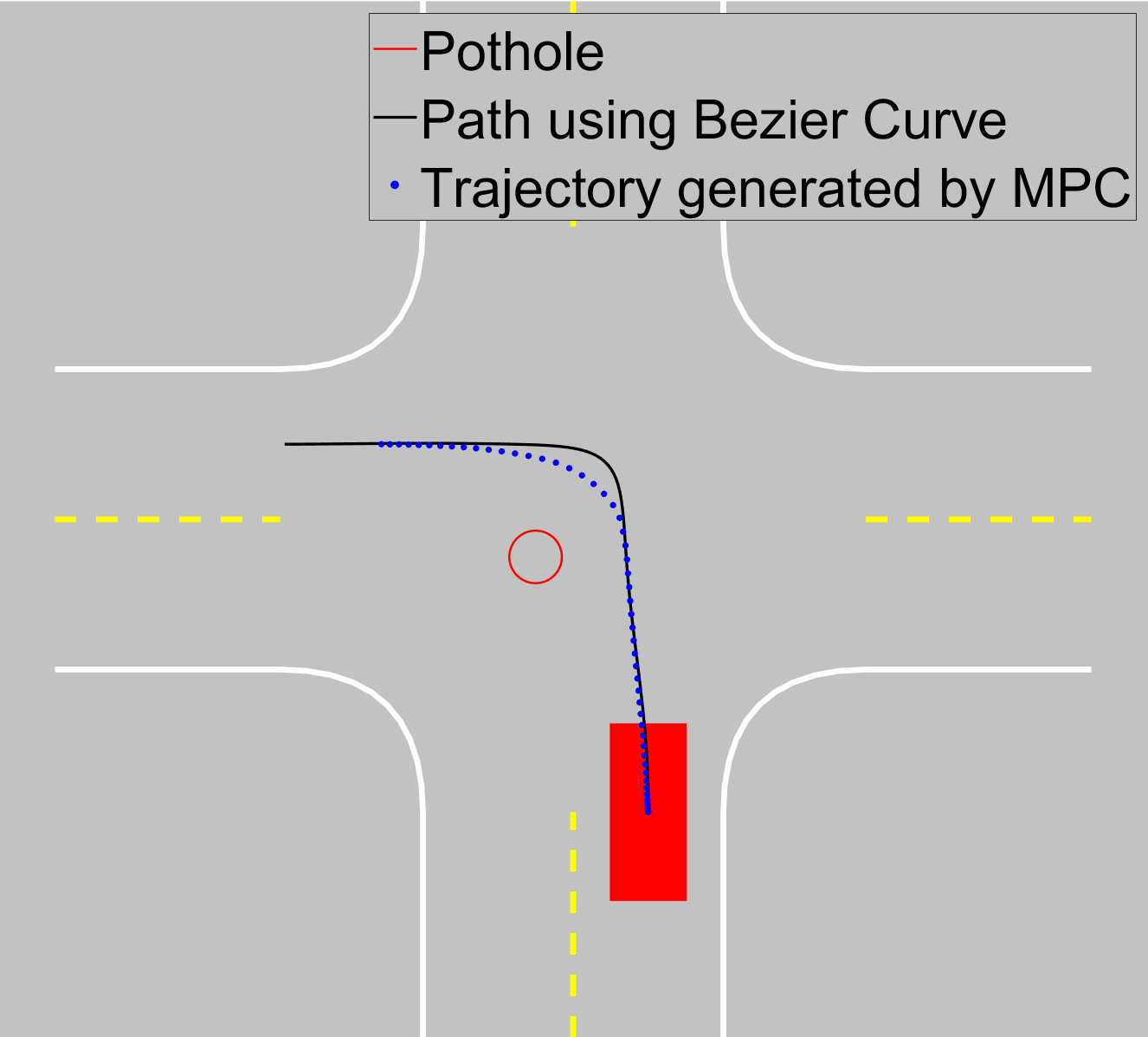}
         \caption{Left turn}
         \label{fig:2a}
     \end{subfigure}
     \begin{subfigure}[b]{0.24\textwidth}
         \centering
         \includegraphics[width=\textwidth]{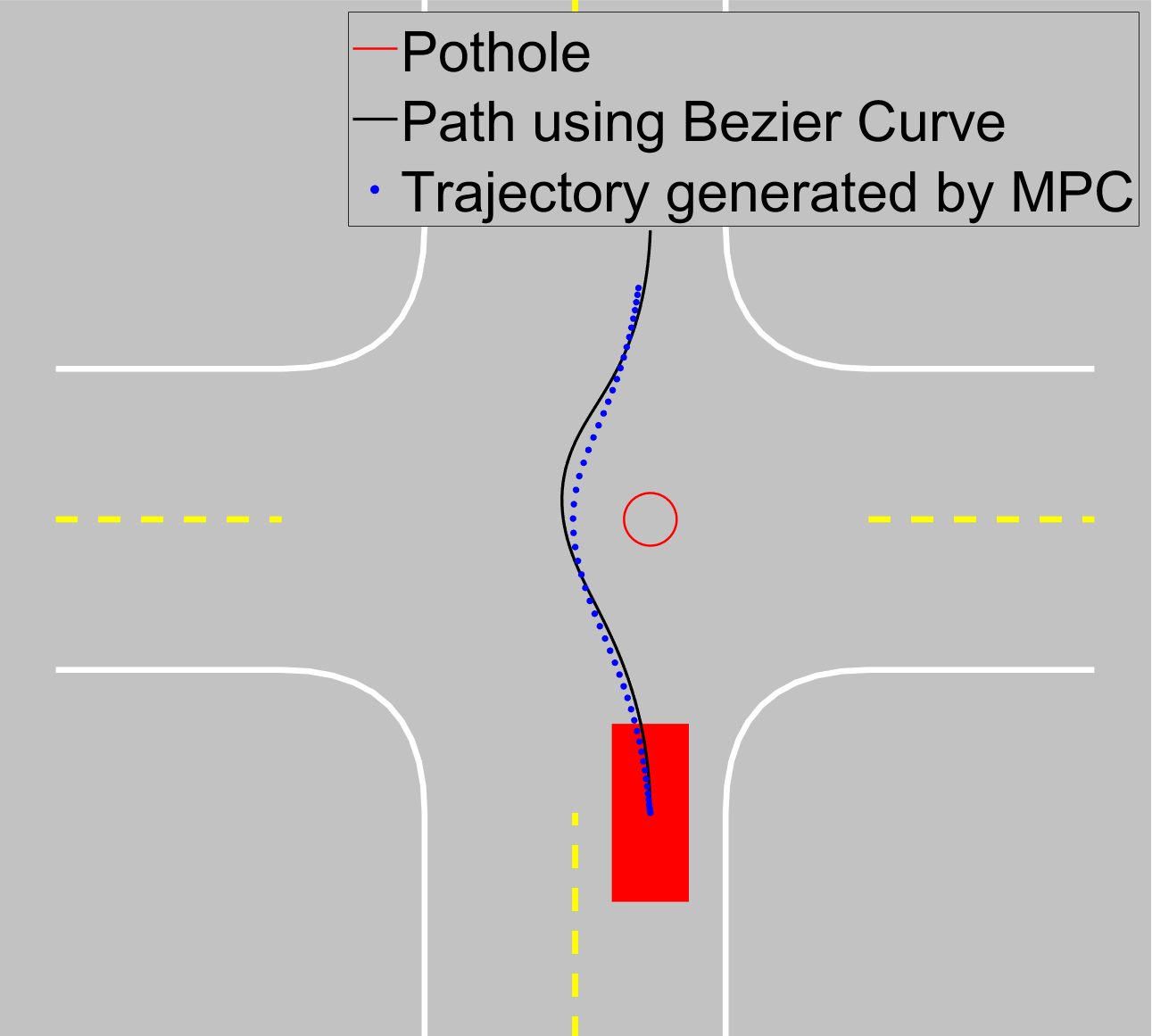}
         \caption{Straight maneuver}
         \label{fig:2b}
     \end{subfigure}
        \caption{Finding a trajectory around a pothole}
        \label{fig:1LPothole}
\end{figure}

The algorithm efficiently finds a trajectory around the potholes as shown in figure \ref{fig:1LPothole}. The trajectory generated by MPC depends on its design, hence an error can be observed in the generated trajectory versus the generated path. For this reason, it is important to have a larger radius of safety for determining the intermediate goal points. In our study, we used the radius equal to the width of the ego vehicle giving an error margin equal to half width of the ego vehicle.

\begin{figure}[htbp]
     \centering
     \begin{subfigure}[b]{0.24\textwidth}
         \centering
         \includegraphics[width=\textwidth]{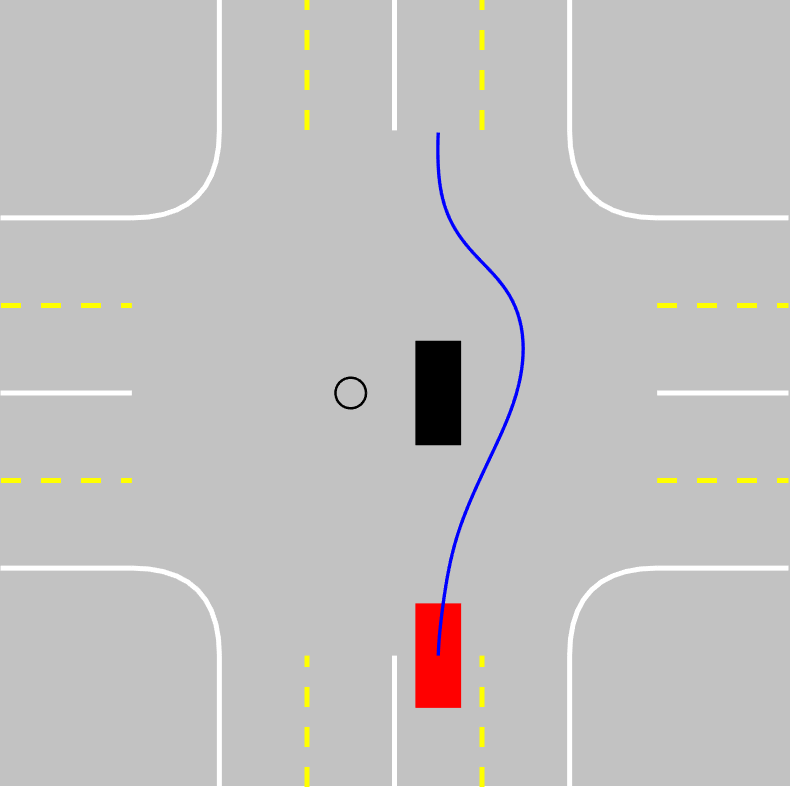}
         \caption{Pothole on the left of the OV}
         \label{fig:3a}
     \end{subfigure}
     \begin{subfigure}[b]{0.24\textwidth}
         \centering
         \includegraphics[width=\textwidth]{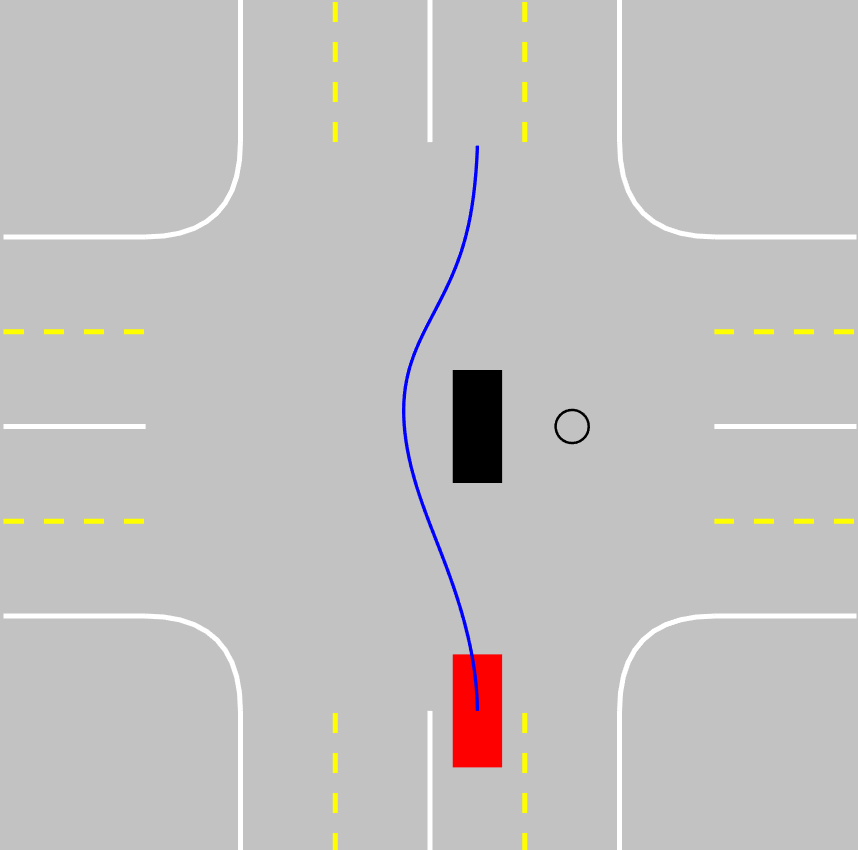}
         \caption{Pothole on the right of the OV}
         \label{fig:3b}
     \end{subfigure}
        \caption{Finding a trajectory around multiple obstacles in a larger environment}
        \label{fig:2LPothole}
\end{figure}
The algorithm can handle multiple obstacles well as seen in figure \ref{fig:2LPothole}. The scenario in the figure shows the case when an obstacle vehicle breaks down in the middle of the intersection and is unable to move. In such cases, it is important for the motion planner to find an alternate route around the vehicle. To make the scenario more complex, potholes were added in the intersection. As seen from the results, the algorithm generates trajectories similar to human intuition.

\begin{figure}[htbp]
     \centering
     \begin{subfigure}[b]{0.24\textwidth}
         \centering
         \includegraphics[width=\textwidth]{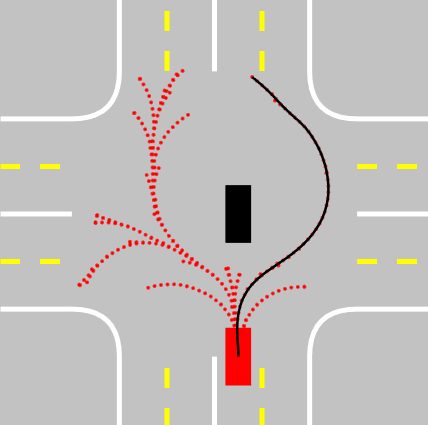}
         \caption{Standard pRRT using $\lambda=10^3$}
         \label{fig:5a}
     \end{subfigure}
     \hfill
     \begin{subfigure}[b]{0.24\textwidth}
         \centering
         \includegraphics[width=\textwidth]{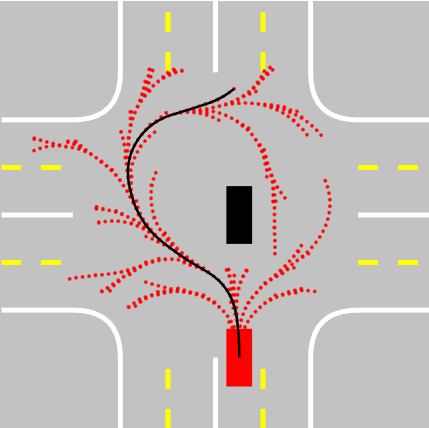}
         \caption{Standard RRT}
         \label{fig:5b}
     \end{subfigure}
     \hfill
     \begin{subfigure}[b]{0.24\textwidth}
         \centering
         \includegraphics[width=\textwidth]{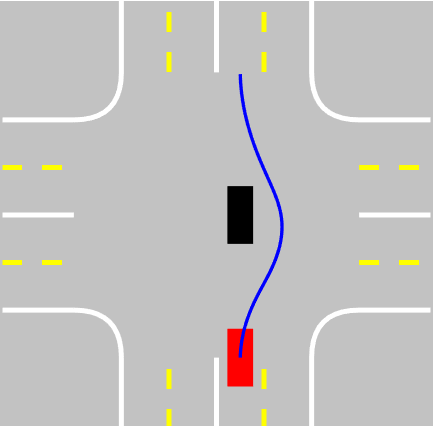}
         \caption{Our method using $\lambda=10^3$}
         \label{fig:5c}
     \end{subfigure}
        \caption{Comparison of our method with other standard methods}
        \label{fig:Comparison}
\end{figure}
Figure \ref{fig:Comparison} shows the performance of our algorithm compared to the standard methods. The trajectories obtained by the standard algorithms are too conservative and do not conform to human intuition when compared to the trajectory obtained by our method. The standard algorithms have higher convergence time as they sample points in unnecessary regions which results in large trees as shown in the figure. This scenario was run 100 times using each of the three techniques to evaluate the performance based on the number of sampling iterations. 
\begin{table}[h!]
\normalsize
\begin{center}
\begin{tabular}{|c|c|}
\hline
& \textbf{Average Iterations}\\ 
\hline
\textbf{RRT} & 925 \\
\hline
\textbf{pRRT} & 637 \\
\hline
\textbf{Our method} & 285 \\
\hline
\end{tabular}
\end{center}
\end{table}

The average number of iterations required for generating the above trajectory successfully is summarized in the table. In the proposed method, the pRRT algorithm is implemented multiple times for the intermediate goals until feasible trajectories are obtained. In the best-case scenario, the first intermediate goal can result in feasible trajectories, and in the worst case, the last intermediate goal can be a successful goal. Hence it is important to reduce the obstacle points of interest which reduces the number of intermediate goals.

\section{Conclusion}
In this paper, we proposed a novel method for the online replanning of autonomous vehicles. The trajectories obtained by the algorithm are safe and in accordance with human intuition. The algorithm was tested in one-lane and two-lane intersection scenarios with potholes and other vehicles occupying the space. Compared with the performance of standard RRT and pRRT algorithms, our method showed superior performance for finding a safe trajectory with less number of iterations. Adding steps to generate trajectories by selecting intermediate goals around the obstacles was found to be more efficient than randomly sampling points in the environment to form a tree from start to destination. To further improve the method, a better MPC tracker can be implemented to track the generated path more accurately. This can effectively reduce the radius of the safety circle and hence resulting in more optimal trajectories. 
For future work, the method will be experimentally implemented in a scaled autonomous vehicle. Experimentally, the ego vehicle will be subjected to more complex scenarios and the algorithm will be evaluated based on the dynamic feasibility, optimality, and closeness to human intuition.
\bibliography{library}          
\end{document}